\documentclass[runningheads]{llncs}
\usepackage[T1]{fontenc}
\usepackage{amssymb}
\usepackage{adjustbox}
\usepackage{tabularx}
\usepackage{ulem}
\usepackage{makecell}
\usepackage{multirow} 
\usepackage{xfp} 

\usepackage{graphicx,verbatim}
\begin{document}
\title{SurgGenesis: A Generative Surgical World Model for Future-Aware Surgical Understanding}
    \titlerunning{SurgGenesis: A Surgical World Model}
    %

\author{
    Qirui Zhong\inst{1} \and
    Quande Liu\inst{2} \and
    Yang Chen\inst{1} \and
    Cheng Xue\inst{1}\textsuperscript{*}}
\authorrunning{Q. Zhong et al.}
\institute{School of Computer Science and Engineering, Southeast University, Nanjing, China\\
\email{220256461@seu.edu.cn}
\and
The Chinese University of Hong Kong (CUHK), Hong Kong SAR, China}
\maketitle              

\begingroup
\renewcommand{\thefootnote}{*}
\footnotetext{Corresponding author: Cheng Xue.}
\endgroup

\begin{abstract}
    Surgical scene understanding is constrained by data scarcity and long-tail class imbalance, as annotating surgical videos demands scarce clinical expertise while rare procedural events remain underrepresented. We tackle this through data generation and propose SurgGenesis, a generative surgical world model that learns future surgical state transitions in a latent space from visual observations and textual procedural descriptions, serving as a scalable surgical data engine. To adapt a large-scale video diffusion model to the data-scarce surgical domain without catastrophic forgetting, we introduce a three-stage progressive LoRA strategy that incrementally injects endoscopic visual priors, procedural semantics, and spatial motion cues. Rather than judging generation by visual fidelity alone, we validate SurgGenesis along two axes that separate a world model from a generic generator: (i) future-awareness, by comparing predicted futures against the real future through a GT-latent prediction gap and temporal-horizon analysis, and (ii) action-faithful usefulness, by checking that generated clips contain the commanded surgical triplet and that using them to rebalance long-tail events and drive downstream understanding (segmentation, triplet recognition, 3D reconstruction) yields action-specific gains. Conditioned futures therefore complement conventional long-tail sampling.\newline Code: \url{https://zh-qr.github.io/SurgGenesis}.


\end{abstract}
\section{Introduction}

\begin{figure}[t]
    \includegraphics[width=\textwidth]{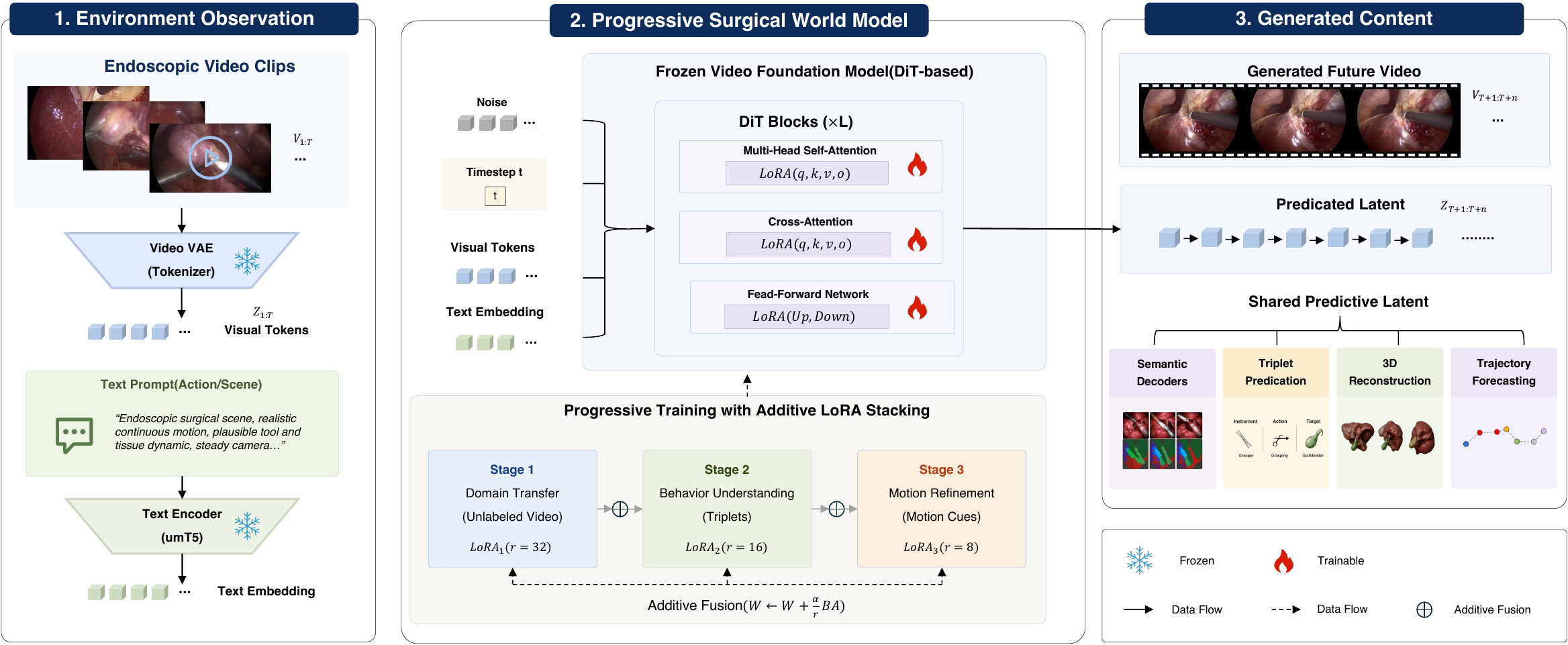}
    \caption{\textbf{Overview of SurgGenesis.} (1) Visual observations and structured procedural text condition the model. (2) A three-stage strategy fine-tunes the foundational video world model. (3) Predicted latent states are decoded into future observations.} \label{Overview}
\end{figure}

Surgical video models can synthesize realistic clips, but visual realism alone is insufficient for a surgical world model. The model must start from an observed operative context, respond to a procedural description, and produce a coherent future. This is demanding in endoscopy, where tool--tissue interactions, viewpoint motion, and procedural phase evolve together.

Representative surgical-vision systems adapt multimodal representations or track instruments from observed videos\cite{walimbe2025adaptation,nwoye2025surgitrack}. Related medical world-model work has modeled latent disease trajectories for prediction and planning\cite{yang2025medical}; SurgGenesis brings this state-transition perspective to text-controlled surgical rollouts.

We use \emph{surgical world model} operationally: given past observations and a structured procedure prompt, the model represents a conditional state transition and rolls out its consequences. A visually plausible sample may still change the operative state or ignore the requested action. The claim therefore requires that the prediction match the real continuation of its context and that the prompt controls the intended action. Generative world models provide a data-driven route to such spatio-temporal modeling\cite{he2025surgworld,jang2025dreamgen,ali2025world,chen2025far,assran2025v,ren2025videoworld}; surgery is a demanding instance\cite{lin2024world,koju2025surgical,ding2025clarity}.

\textbf{SurgGenesis} models the conditional future transition $p(\hat{Z}_{t+1:T+n}\mid Z_{1:T},c)$, where $c$ is a structured textual description of procedure phase and instrument--verb--target (IVT) state. Unlike a generator assessed only by clip realism\cite{chen2025surgsora,sun2024bora,wang2025feat}, SurgGenesis is evaluated as a context-conditioned, text-controllable rollout model. A \textbf{Three-Stage Progressive LoRA Fine-Tuning Strategy} adapts a large video diffusion model to endoscopic appearance, procedural semantics, and spatial dynamics without catastrophic forgetting. Additive LoRA stacking and rehearsal preserve earlier capabilities while later stages specialize the state-transition model.

Controlled rollouts can create rare, context-specific examples for costly, long-tailed surgical datasets\cite{li2024endora,murali2023endoscapes,hong2020cholecseg8k}. Data scarcity and imbalance are application settings, not the sole claim: synthetic clips are useful only when they preserve the supplied context and requested action.

Our evaluation separates model properties from application utility. Fidelity metrics quantify visual realism. The Prediction Gap and multi-horizon RMSE test whether a rollout matches its real continuation. Thus, future awareness concerns state transitions rather than data scarcity. A real-data recognizer tests whether the commanded IVT appears in the generated future. Segmentation, IVT recognition, 3D reconstruction, and long-tail augmentation then assess utility. Thus, a downstream gain or FVD improvement alone is not presented as evidence of a world model.

The main contributions of this paper are summarized as follows:
\begin{itemize}
    \item We introduce \textbf{SurgGenesis}, a text-controllable generative surgical world model that rolls out future videos from observed surgical context and structured prompts containing procedure phase and IVT state.
    
    \item We propose a three-stage progressive LoRA adaptation strategy with additive stacking and data rehearsal, progressively incorporating endoscopic visual priors, procedural semantics, and spatial dynamics while mitigating catastrophic forgetting.
    
    \item We introduce a layered evaluation: fidelity metrics assess visual quality; the Prediction Gap and temporal horizons test context-to-future consistency; an IVT recognizer tests action faithfulness; and downstream tasks assess application utility.
\end{itemize}

\section{Methodology}

An overview of the proposed SurgGenesis framework is shown in Fig.~\ref{Overview}.
SurgGenesis learns predictive representations of surgical dynamics by integrating visual observations with structured procedural descriptions in a shared latent space. The resulting model rolls out a future surgical state that is conditioned on both the current scene and a user-specified procedure state.

\subsection{Problem Formulation}

We formulate surgical scene forecasting as the prediction of future latent states conditioned on historical observations and a structured prompt.
Given a sequence of $T$ past frames, denoted as $V_{1:T}$, the model predicts the next $n$ latent states $\hat{Z}_{T+1:T+n}$ rather than generating pixels directly. Formally, it learns the conditional future distribution $p(\hat{Z}_{T+1:T+n}\mid Z_{1:T},c)$, where $c$ contains procedural text and coordinate-aware motion cues, and generates the future latent clip jointly through conditional flow matching.
The text prompt is deliberately structured so that the phase, instrument, verb, and target can be changed independently: ``Endoscopic surgical scene. Laparoscopic cholecystectomy, phase: \{\textit{phase}\}. At frame 0, the \{\textit{instrument}\} is \{\textit{verb}\} the \{\textit{target}\}. Predict the next 60 frames with smooth instrument motion, stable endoscopic viewpoint, and anatomically consistent tool--tissue interaction.'' For example, the fields can specify gallbladder extraction, a grasper, retracting, and the cystic plate. This representation turns the procedure description into an intervention-like control signal while retaining the observed scene as state context.
The prompt controls the intended procedural transition, whereas the input frames anchor the anatomy, tool configuration, illumination, and endoscopic viewpoint. Its role is therefore not unconstrained free-form synthesis: phase constrains which IVT combinations are meaningful, and the generated continuation must remain compatible with the visual state at frame~0. This separation is central to the proposed world-model setting. A generic text-to-video generator may satisfy the textual description while changing the scene arbitrarily; SurgGenesis must preserve the scene state while rolling out the requested action.
We operationalize future awareness as context-to-future consistency: predicted latents and decoded frames should match the corresponding real continuation, not merely resemble surgery. Direct latent RMSE, decoded frame-space RMSE, and the RT-DETR trajectory readout are distinct diagnostics. The representation is therefore trained to capture spatio-temporal regularities such as instrument geometry, viewpoint continuity, and tool--tissue motion; segmentation and multi-view 3D reconstruction probe whether these rollouts preserve structural and geometric information.

\subsection{Predictive Surgical World Modeling via Progressive Training}

Surgical procedures evolve as complex and dynamic processes. However, most existing approaches focus on static interpretation of observed frames, which limits their ability to model future surgical states. To address this limitation, we develop a predictive surgical world model through \textbf{Progressive Surgical Dynamics Training}, a multi-stage paradigm designed to progressively incorporate domain-specific knowledge into a latent diffusion framework. The core of the model lies in the high-dimensional latent space $Z$ introduced above, where a Diffusion Transformer (DiT) learns predictive representations of surgical dynamics by modeling the temporal evolution of surgical scenes. Following DiffSynth-Studio, training uses its native flow-matching SFT objective rather than direct $L_1$ latent regression: $\mathcal{L}_{\mathrm{FM}}=\mathbb{E}_{Z_0,\epsilon,t}[w(t)\|v_\theta(Z_t,t,c)-v_t^\star\|_2^2]$, where the scheduler constructs noisy latent $Z_t$ and flow target $v_t^\star$ from clean future latent $Z_0$, Gaussian noise $\epsilon$, and sampled timestep $t$. The first-frame latent is clamped as conditioning and excluded from the loss.

\textbf{Stage 1: Domain Transfer.} The model is initially adapted to the surgical domain using large-scale unlabeled video clips (Cholec80) and fixed textual descriptions.
Low-Rank Adaptation (LoRA) is applied across all 30 transformer blocks to effectively capture surgical visual priors, such as specific tissue textures and illumination, without requiring dense manual annotations.

\textbf{Stage 2: Behavior Understanding.} Procedural knowledge is injected by aligning visual transitions with structured prompts containing phase and instrument--verb--target fields. These fields can be varied at inference time while the observed frames retain the operative context.
To prevent the catastrophic forgetting of Stage 1 capabilities, we introduce \textit{Additive LoRA Stacking}: the learned LoRA weights from Stage 1 are permanently fused into the base model weights via calculated increments.
A new LoRA is then initialized and restricted exclusively to the last 20 transformer blocks, reducing plasticity to protect early-layer feature extraction.
Furthermore, a 20\% data rehearsal strategy is employed, mixing baseline dataset samples into the current training batches to maintain foundational stability.

\textbf{Stage 3: Spatial Dynamics Refinement.} We introduce coordinate-aware descriptions of instrument motion to improve spatial precision and physical plausibility.
Following the additive stacking principle, Stage 2 weights are fused, and the final LoRA targets only the deepest 10 blocks with a highly conservative learning rate.
This ensures the model learns fine-grained spatial trajectories without disrupting the harmonized text distribution and procedural semantics acquired in previous stages.

\subsection{Layered Evaluation and Downstream Verification}

Our evaluation separates controlled world-model behavior from application utility. The Prediction Gap isolates forecasting error from decoder capacity by comparing predicted-future decoding against an upper bound from Ground-Truth (GT) latents. FVD-style and LPIPS metrics characterize visual or distributional fidelity; the Prediction Gap and temporal horizons characterize context-to-future consistency; and the IVT recognizer characterizes prompt-level action faithfulness. Downstream tasks then test whether the generated rollouts are useful, rather than treating any single downstream gain as proof of a world model.

The application-level protocol evaluates augmentation, segmentation, triplet recognition, and 3D reconstruction. These tasks probe utility, structure, semantics, and geometry.

\section{Experiments}
\subsection{Datasets}

    \begin{figure}[t]
        \includegraphics[width=\textwidth]{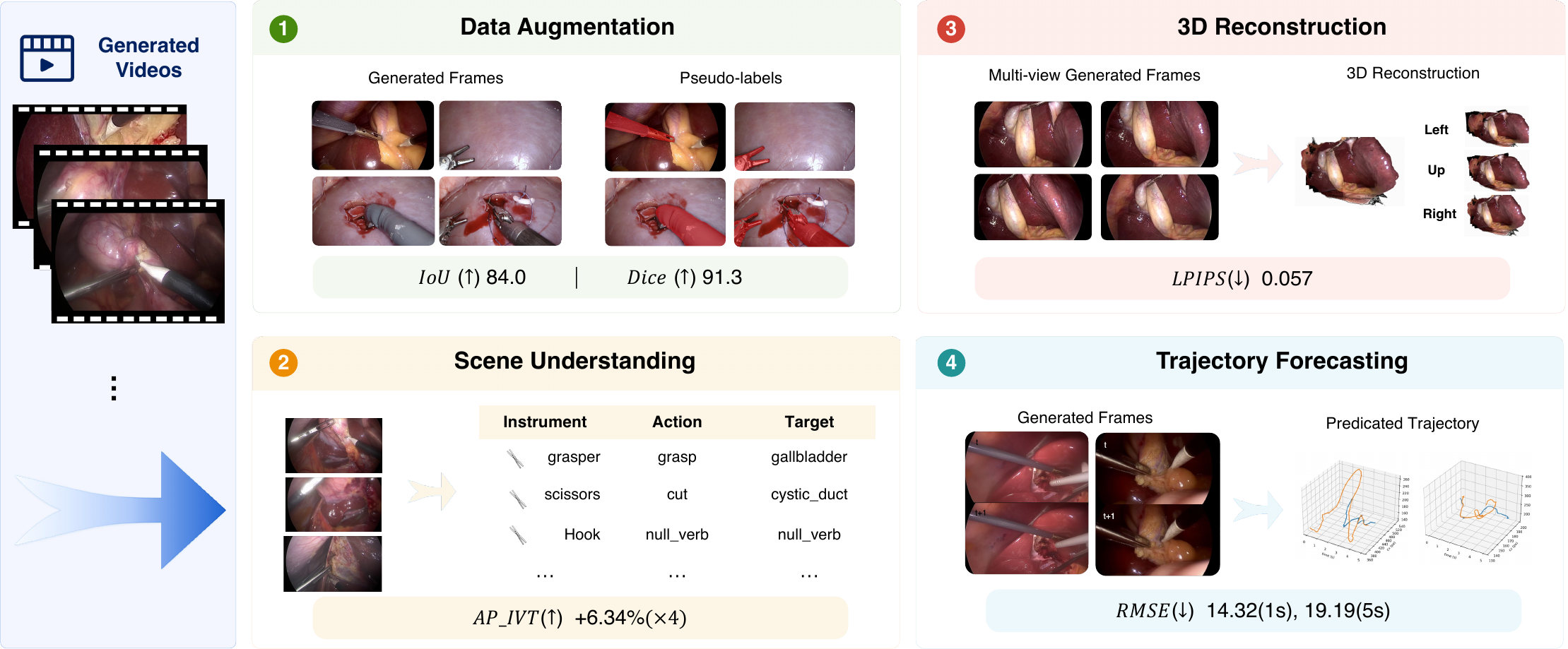}
        \caption{\textbf{Application-level utility of SurgGenesis.} (1) Data augmentation. (2) Scene understanding. (3) 3D reconstruction. (4) Future-trajectory readout, evaluated by aligned frame-space RMSE; RMSE is multiplied by 100 for readability.} \label{fig:downstream-evidence}
    \end{figure}

Experiments use the Cholec-series datasets. Cholec80\cite{twinanda2016endonet}, CholecT50\cite{nwoye2022rendezvous}, and CholecTrack20\cite{nwoye2025cholectrack20} support Stages~1--3, respectively. Segmentation uses CholecSeg8k\cite{hong2020cholecseg8k} and Endoscapes2023\cite{murali2023endoscapes}; triplet recognition and augmentation use CholecT50; and 3D reconstruction uses EndoNeRF\cite{wang2022neural}. All downstream benchmarks follow official partitions and standard protocols. Generation and internal evaluation use held-out clips with active instrument use. In the official five-fold CholecT50 protocol, conditioning, synthesis, and downstream training use only each fold's training partition; validation and test remain unchanged.

\subsection{Implementation Details}

All experiments are conducted on NVIDIA A100 GPUs. We adopt the dense WAN2.2-TI2V-5B\cite{wan2025wan} as the foundational world model and fine-tune it with progressive LoRA in DiffSynth-Studio. Its unified text/image-to-video formulation and high-compression WAN2.2-VAE provide the backbone for surgical future generation.
The progressive LoRA stacking uses ranks 32/16/8 over the last 30/20/10 blocks for Stages 1--3, with learning rates 5e-5/5e-6/2e-6 and a 20\% Cholec80 rehearsal ratio in Stages 2--3. Standard clips are generated at $832\times480$ in BF16 using 41 conditioning and 40 predicted frames. The temporal-horizon benchmark separately generates 81-frame future clips at 15 FPS, covering 5.33\,s; the 1--5\,s metrics use frame offsets 15/30/45/60/75, and a horizon is omitted when a clip does not provide sufficient coverage.
Complementing the DiT as the forward state-transition model, an RT-DETR-based inverse dynamics model\cite{zhao2024detrs} recovers instrument actions from predicted frames and provides the trajectory readout. Controllability is judged separately by a CholecT50 recognizer trained only on real data. For the stagewise feature probe, we duplicate the final two WAN-VAE layers, extract their features, and apply a convolutional classifier; this lightweight head is fixed across stages to isolate representation changes. For segmentation, we append a lightweight decoder-side convolutional head that maps three channels to one mask channel and retrain the decoder/head, relying on the foundation model's surgical representation rather than a SOTA segmentation architecture. For the main IVT comparison, we reuse Rendezvous\cite{nwoye2022rendezvous} solely as the triplet-recognition head; it is not used to generate segmentation masks. A separate fixed SAM-based evaluator measures pseudo-mask overlap between generated and real future frames. During downstream evaluation, the DiT remains frozen and only task heads are trained.

\subsection{Ablation Study}

We evaluate the three-stage strategy on generation fidelity and context-to-future consistency. The R3D-18 FVD-style score measures video fidelity, while frame-space RMSE compares generated frames with aligned real futures over 1--5s horizons. It is distinct from direct latent RMSE and the RT-DETR trajectory readout. Fig.~\ref{fig:quant-ablation} reports full stagewise curves: all adapted stages improve over native WAN in FVD-style fidelity, while Stage~2 and Stage~3 are closely matched at 5s. Table~\ref{tab:downstream-validation} summarizes native-WAN-to-Stage-3 results. Fig.~\ref{fig:quant-ablation}(a) and the Prediction Gap use native $[0,1]$ RMSE; Fig.~\ref{fig:downstream-evidence} and Table~\ref{tab:downstream-validation} report $100\times$ frame RMSE.

\subsection{Downstream Tasks}

\begin{figure}
    \centering
    \includegraphics[width=\textwidth]{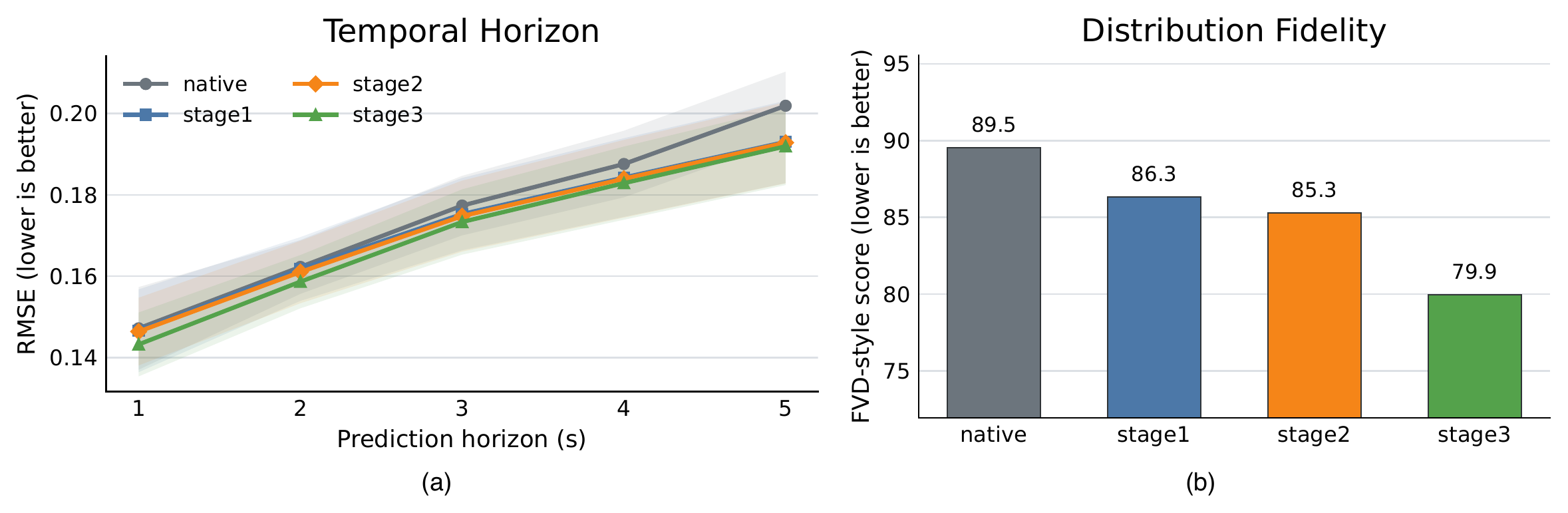}
    \caption{\textbf{Ablation and results.} (a) Temporal-horizon frame RMSE on the native $[0,1]$ scale. (b) Stage-wise R3D-18 FVD-style score.}
    \label{fig:quant-ablation}
\end{figure}

We evaluate controlled surgical rollouts before asking whether they are useful downstream. Table~\ref{tab:downstream-results} summarizes reference results for generation fidelity (LPIPS), rare-class augmentation (triplet $mAP_{IVT}$, following CholecT50\cite{nwoye2022rendezvous}), and 3D reconstruction (LPIPS on VGGT4D\cite{hu2025vggt4d} views). Because conditioning and protocols differ, the table provides context rather than a controlled head-to-head ranking. Distribution-level fidelity is reported in Fig.~\ref{fig:quant-ablation}. We do not report FVD against unconditional generators such as Endora\cite{li2024endora}: they synthesize from noise, whereas SurgGenesis rolls out a future from given frames and a structured prompt.

\textbf{State-Transition Consistency: Predicted vs.\ Real Future.}
Visual fidelity does not establish that a model preserves the observed state. We therefore compare each rollout with its aligned real continuation. The Prediction Gap is computed in decoded frame space: GT latents give an upper-bound frame RMSE@5s of $0.0227$, whereas Stage-3 predicted latents give $0.1888$, for a gap of $0.1661$. The dominant residual is therefore attributable to future prediction rather than VAE decoding. Direct latent RMSE is a separate diagnostic; the 1--5s curve reports aligned frame RMSE, and RT-DETR provides a complementary trajectory readout. This supplies state-transition evidence for the world-model formulation, not a claim about augmentation.

\textbf{Text Controllability.}
An independent ConvNeXt-Tiny CholecT50 recognizer trained only on real data tests whether Stage-3 futures contain the IVT specified by the prompt. Across 432 clips and 24 minority IVT classes, it recovers the commanded triplet in \textbf{93.06\%} of cases (instrument 97.45\%, verb 90.97\%, target 93.06\%; five-fold range 87.5--96.6\%), versus 6.25\% for mismatched triplets. Mismatched prompts still recover shared instrument/verb components above 74\%; target and full-triplet recovery are thus the more discriminative control tests. Together with state-transition consistency, this result supports context-grounded generation of the requested procedural state.

\begin{table}[t]
    \caption{\textbf{Generation on future frames.} Reference results for fidelity (LPIPS$\downarrow$), rare-class triplet augmentation ($mAP_{IVT}\uparrow$), and 3D reconstruction (LPIPS$\downarrow$). Protocols differ; bold marks the best reported value, not a controlled ranking.}\label{tab:downstream-results}
    \centering
    \footnotesize
    \renewcommand{\arraystretch}{1.0}
    \setlength{\tabcolsep}{8pt}
    \begin{adjustbox}{max width=\linewidth}
    \begin{tabular}{lccc}
    \hline
    \textbf{Method} & \textbf{Fidelity (LPIPS)}$\downarrow$ & \textbf{Triplet ($mAP_{IVT}$)}$\uparrow$ & \textbf{3D (LPIPS)}$\downarrow$ \\
    \hline
    SimDA\cite{xing2024simda}    & 0.565 & 32.7 & 0.089 \\
    VDM\cite{ho2022video}        & 0.652 & 32.9 & 0.074 \\
    VidGPT\cite{yan2021videogpt} & 0.563 & 35.8 & 0.089 \\
    EndoGen\cite{liu2025endogen} & 0.528 & 36.5 & \textbf{0.047} \\
    \textbf{SurgGenesis}         & \textbf{0.516} & \textbf{36.9} & 0.057 \\
    \hline
    \end{tabular}
    \end{adjustbox}
\end{table}

\begin{table}[t]
    \caption{\textbf{Stagewise validation.} The baseline is native WAN except for $AP_{IVT}$, which uses the real-only classifier. Native WAN failed under the 3D reconstruction pipeline, so its LPIPS is unavailable.}\label{tab:downstream-validation}
    \centering
    \footnotesize
    \renewcommand{\arraystretch}{1.0}
    \setlength{\tabcolsep}{6pt}
    \begin{adjustbox}{max width=\linewidth}
    \begin{tabular}{llccc}
    \hline
    \textbf{Downstream task} & \textbf{Metric} & \textbf{Baseline} & \textbf{Stage-3} & \textbf{Gain} \\
    \hline
    Future trajectory & Frame RMSE@5s ($\times100$) $\downarrow$ & 20.19  & \textbf{19.19}  & $-4.94\%$ \\
    3D reconstruction & LPIPS $\downarrow$            & Failed & \textbf{0.057}  & -- \\
    SAM pseudo-mask probe & IoU / Dice $\uparrow$      & 0.2494 / 0.3117 & \textbf{0.2615 / 0.3221} & $+4.85 / +3.34\%$ \\
    Video fidelity & R3D-18 FVD-style $\downarrow$    & 89.54  & \textbf{79.94}  & $-10.72\%$ \\
    Long-tail recognition & $AP_{IVT}\uparrow$ ($\times4$) & 0.1717 & \textbf{0.1823} & $+6.34\%$ \\
    \hline
    \end{tabular}
    \end{adjustbox}
\end{table}

\noindent\emph{Application utility.} Long-tail augmentation is an application-level test, not the definition of the world model. At matched $4\times$ scale, SurgGenesis reaches 0.1823 $AP_{IVT}$, versus 0.1815 for Copy$\times$4; rebalancing explains most of the gain, while synthesis adds modest action-conditioned variation. Under the non-identical protocols of Table~\ref{tab:downstream-results}, SurgGenesis is competitive in fidelity, triplet augmentation, and 3D reconstruction. The decoder-retrained segmentation head reaches $84.0$ IoU / $91.3$ Dice; Rendezvous is used only for IVT recognition. The fixed SAM diagnostic improves over native WAN (IoU 0.2615 vs.\ 0.2494; Dice 0.3221 vs.\ 0.3117), indicating useful action-specific variation without attributing the entire augmentation gain to synthesis.

\noindent\emph{Interpreting the evidence.} FVD-style and LPIPS assess fidelity; the Prediction Gap and horizon-wise RMSE assess context-to-future consistency; and IVT recovery assesses prompt compliance. Segmentation, 3D reconstruction, and long-tail augmentation then test retained structure and downstream utility. The world-model claim rests on the first three levels together.

\section{Conclusion}

SurgGenesis is a text-controllable world model that rolls out futures from observed operative states and structured prompts. Fidelity, state-transition consistency, and IVT recovery assess its rollouts; downstream tasks test their utility.

    \newpage

    \bibliographystyle{splncs04}
    \bibliography{reference}

@article{jang2025dreamgen,
  title={Dreamgen: Unlocking generalization in robot learning through video world models},
  author={Jang, Joel and Ye, Seonghyeon and Lin, Zongyu and Xiang, Jiannan and Bjorck, Johan and Fang, Yu and Hu, Fengyuan and Huang, Spencer and Kundalia, Kaushil and Lin, Yen-Chen and others},
  journal={arXiv preprint arXiv:2505.12705},
  year={2025}
}

@article{he2025surgworld,
  title={SurgWorld: Learning Surgical Robot Policies from Videos via World Modeling},
  author={He, Yufan and Guo, Pengfei and Xu, Mengya and Li, Zhaoshuo and Myronenko, Andriy and Imans, Dillan and Liu, Bingjie and Yang, Dongren and Gu, Mingxue and Ji, Yongnan and others},
  journal={arXiv preprint arXiv:2512.23162},
  year={2025}
}

@article{lin2024world,
  title={World models for general surgical grasping},
  author={Lin, Hongbin and Li, Bin and Wong, Chun Wai and Rojas, Juan and Chu, Xiangyu and Au, Kwok Wai Samuel},
  journal={arXiv preprint arXiv:2405.17940},
  year={2024}
}

@article{assran2025v,
  title={V-jepa 2: Self-supervised video models enable understanding, prediction and planning},
  author={Assran, Mido and Bardes, Adrien and Fan, David and Garrido, Quentin and Howes, Russell and Muckley, Matthew and Rizvi, Ammar and Roberts, Claire and Sinha, Koustuv and Zholus, Artem and others},
  journal={arXiv preprint arXiv:2506.09985},
  year={2025}
}

@article{ali2025world,
  title={World simulation with video foundation models for physical ai},
  author={Ali, Arslan and Bai, Junjie and Bala, Maciej and Balaji, Yogesh and Blakeman, Aaron and Cai, Tiffany and Cao, Jiaxin and Cao, Tianshi and Cha, Elizabeth and Chao, Yu-Wei and others},
  journal={arXiv preprint arXiv:2511.00062},
  year={2025}
}

@article{chen2025far,
  title={How Far Are Surgeons from Surgical World Models? A Pilot Study on Zero-shot Surgical Video Generation with Expert Assessment},
  author={Chen, Zhen and Xu, Qing and Wu, Jinlin and Yang, Biao and Zhai, Yuhao and Guo, Geng and Zhang, Jing and Ding, Yinlu and Navab, Nassir and Luo, Jiebo},
  journal={arXiv preprint arXiv:2511.01775},
  year={2025}
}

@article{ding2025clarity,
  title={CLARITY: Medical World Model for Guiding Treatment Decisions by Modeling Context-Aware Disease Trajectories in Latent Space},
  author={Ding, Tianxingjian and Zou, Yuanhao and Chen, Chen and Shah, Mubarak and Tian, Yu},
  journal={arXiv preprint arXiv:2512.08029},
  year={2025}
}

@article{yang2025medical,
  title={Medical world model: Generative simulation of tumor evolution for treatment planning},
  author={Yang, Yijun and Wang, Zhao-Yang and Liu, Qiuping and Sun, Shuwen and Wang, Kang and Chellappa, Rama and Zhou, Zongwei and Yuille, Alan and Zhu, Lei and Zhang, Yu-Dong and others},
  journal={arXiv preprint arXiv:2506.02327},
  year={2025}
}

@inproceedings{wang2025feat,
  title={FEAT: Full-Dimensional Efficient Attention Transformer for Medical Video Generation},
  author={Wang, Huihan and Yang, Zhiwen and Zhang, Hui and Zhao, Dan and Wei, Bingzheng and Xu, Yan},
  booktitle={International Conference on Medical Image Computing and Computer-Assisted Intervention},
  pages={267--277},
  year={2025},
  organization={Springer}
}

@inproceedings{li2024endora,
  title={Endora: Video generation models as endoscopy simulators},
  author={Li, Chenxin and Liu, Hengyu and Liu, Yifan and Feng, Brandon Y and Li, Wuyang and Liu, Xinyu and Chen, Zhen and Shao, Jing and Yuan, Yixuan},
  booktitle={International conference on medical image computing and computer-assisted intervention},
  pages={230--240},
  year={2024},
  organization={Springer}
}

@inproceedings{koju2025surgical,
  title={Surgical vision world model},
  author={Koju, Saurabh and Bastola, Saurav and Shrestha, Prashant and Amgain, Sanskar and Shrestha, Yash Raj and Poudel, Rudra PK and Bhattarai, Binod},
  booktitle={MICCAI Workshop on Data Engineering in Medical Imaging},
  pages={1--10},
  year={2025},
  organization={Springer}
}

@inproceedings{liu2025endogen,
  title={Endogen: Conditional autoregressive endoscopic video generation},
  author={Liu, Xinyu and Liu, Hengyu and Wang, Cheng and Liu, Tianming and Yuan, Yixuan},
  booktitle={International Conference on Medical Image Computing and Computer-Assisted Intervention},
  pages={169--179},
  year={2025},
  organization={Springer}
}

@article{twinanda2016endonet,
  title={Endonet: a deep architecture for recognition tasks on laparoscopic videos},
  author={Twinanda, Andru P and Shehata, Sherif and Mutter, Didier and Marescaux, Jacques and De Mathelin, Michel and Padoy, Nicolas},
  journal={IEEE transactions on medical imaging},
  volume={36},
  number={1},
  pages={86--97},
  year={2016},
  publisher={IEEE}
}

@article{nwoye2022rendezvous,
  title={Rendezvous: Attention mechanisms for the recognition of surgical action triplets in endoscopic videos},
  author={Nwoye, Chinedu Innocent and Yu, Tong and Gonzalez, Cristians and Seeliger, Barbara and Mascagni, Pietro and Mutter, Didier and Marescaux, Jacques and Padoy, Nicolas},
  journal={Medical Image Analysis},
  volume={78},
  pages={102433},
  year={2022},
  publisher={Elsevier}
}

@inproceedings{nwoye2025cholectrack20,
  title={Cholectrack20: A multi-perspective tracking dataset for surgical tools},
  author={Nwoye, Chinedu Innocent and Elgohary, Kareem and Srinivas, Anvita and Zaid, Fauzan and Lavanchy, Jo{\"e}l L and Padoy, Nicolas},
  booktitle={Proceedings of the IEEE/CVF Conference on Computer Vision and Pattern Recognition},
  pages={8942--8952},
  year={2025}
}

@article{hong2020cholecseg8k,
  title={Cholecseg8k: a semantic segmentation dataset for laparoscopic cholecystectomy based on cholec80},
  author={Hong, W-Y and Kao, C-L and Kuo, Y-H and Wang, J-R and Chang, W-L and Shih, C-S},
  journal={arXiv preprint arXiv:2012.12453},
  year={2020}
}

@article{murali2023endoscapes,
  title={The endoscapes dataset for surgical scene segmentation, object detection, and critical view of safety assessment: Official splits and benchmark},
  author={Murali, Aditya and Alapatt, Deepak and Mascagni, Pietro and Vardazaryan, Armine and Garcia, Alain and Okamoto, Nariaki and Costamagna, Guido and Mutter, Didier and Marescaux, Jacques and Dallemagne, Bernard and others},
  journal={arXiv preprint arXiv:2312.12429},
  year={2023}
}

@inproceedings{walimbe2025adaptation,
  title={Adaptation of Multi-modal Representation Models for Multi-task Surgical Computer Vision},
  author={Walimbe, Soham and Baby, Britty and Srivastav, Vinkle and Padoy, Nicolas},
  booktitle={International Conference on Medical Image Computing and Computer-Assisted Intervention},
  pages={24--33},
  year={2025},
  organization={Springer}
}

@article{nwoye2025surgitrack,
  title={SurgiTrack: Fine-grained multi-class multi-tool tracking in surgical videos},
  author={Nwoye, Chinedu Innocent and Padoy, Nicolas},
  journal={Medical Image Analysis},
  volume={101},
  pages={103438},
  year={2025},
  publisher={Elsevier}
}

@article{wan2025wan,
  title={Wan: Open and advanced large-scale video generative models},
  author={Wan, Team and Wang, Ang and Ai, Baole and Wen, Bin and Mao, Chaojie and Xie, Chen-Wei and Chen, Di and Yu, Feiwu and Zhao, Haiming and Yang, Jianxiao and others},
  journal={arXiv preprint arXiv:2503.20314},
  year={2025}
}

@inproceedings{ren2025videoworld,
  title={Videoworld: Exploring knowledge learning from unlabeled videos},
  author={Ren, Zhongwei and Wei, Yunchao and Guo, Xun and Zhao, Yao and Kang, Bingyi and Feng, Jiashi and Jin, Xiaojie},
  booktitle={Proceedings of the Computer Vision and Pattern Recognition Conference},
  pages={29029--29039},
  year={2025}
}

@article{hu2025vggt4d,
  title={VGGT4D: Mining Motion Cues in Visual Geometry Transformers for 4D Scene Reconstruction},
  author={Hu, Yu and Cheng, Chong and Yu, Sicheng and Guo, Xiaoyang and Wang, Hao},
  journal={arXiv preprint arXiv:2511.19971},
  year={2025}
}

@inproceedings{xing2024simda,
  title={Simda: Simple diffusion adapter for efficient video generation},
  author={Xing, Zhen and Dai, Qi and Hu, Han and Wu, Zuxuan and Jiang, Yu-Gang},
  booktitle={Proceedings of the IEEE/CVF conference on computer vision and pattern recognition},
  pages={7827--7839},
  year={2024}
}

@article{ho2022video,
  title={Video diffusion models},
  author={Ho, Jonathan and Salimans, Tim and Gritsenko, Alexey and Chan, William and Norouzi, Mohammad and Fleet, David J},
  journal={Advances in neural information processing systems},
  volume={35},
  pages={8633--8646},
  year={2022}
}

@article{yan2021videogpt,
  title={Videogpt: Video generation using vq-vae and transformers},
  author={Yan, Wilson and Zhang, Yunzhi and Abbeel, Pieter and Srinivas, Aravind},
  journal={arXiv preprint arXiv:2104.10157},
  year={2021}
}

@inproceedings{wang2022neural,
  title={Neural rendering for stereo 3d reconstruction of deformable tissues in robotic surgery},
  author={Wang, Yuehao and Long, Yonghao and Fan, Siu Hin and Dou, Qi},
  booktitle={International conference on medical image computing and computer-assisted intervention},
  pages={431--441},
  year={2022},
  organization={Springer}
}

@article{sun2024bora,
  title={Bora: biomedical generalist video generation model},
  author={Sun, Weixiang and You, Xiaocao and Zheng, Ruizhe and Yuan, Zhengqing and Li, Xiang and He, Lifang and Li, Quanzheng and Sun, Lichao},
  journal={arXiv preprint arXiv:2407.08944},
  year={2024}
}

@inproceedings{chen2025surgsora,
  title={Surgsora: Object-aware diffusion model for controllable surgical video generation},
  author={Chen, Tong and Yang, Shuya and Wang, Junyi and Bai, Long and Ren, Hongliang and Zhou, Luping},
  booktitle={International Conference on Medical Image Computing and Computer-Assisted Intervention},
  pages={521--531},
  year={2025},
  organization={Springer}
}

@inproceedings{zhao2024detrs,
  title={Detrs beat yolos on real-time object detection},
  author={Zhao, Yian and Lv, Wenyu and Xu, Shangliang and Wei, Jinman and Wang, Guanzhong and Dang, Qingqing and Liu, Yi and Chen, Jie},
  booktitle={Proceedings of the IEEE/CVF conference on computer vision and pattern recognition},
  pages={16965--16974},
  year={2024}
}

    \end{document}